\documentclass[journal]{IEEEtran}
\usepackage[utf8]{inputenc}
\usepackage[T1]{fontenc}
\usepackage{url}
\usepackage{graphicx}
\usepackage{amsmath}

\newenvironment{changemargin}[2]{\begin{list}{}{%
\setlength{\topsep}{0pt}%
\setlength{\leftmargin}{0pt}%
\setlength{\rightmargin}{0pt}%
\setlength{\listparindent}{\parindent}%
\setlength{\itemindent}{\parindent}%
\setlength{\parsep}{0pt plus 1pt}%
\addtolength{\leftmargin}{#1}%
\addtolength{\rightmargin}{#2}%
}\item }{\end{list}}

\ifCLASSINFOpdf
\else
\fi
\begin{document}
%
% paper title
% can use linebreaks \\ within to get better formatting as desired
\title{Agent-Oriented Approach for Detecting and Managing 
Risks in Emergency Situations}
%
%
% author names and IEEE memberships
% note positions of commas and nonbreaking spaces ( ~ ) LaTeX will not break
% a structure at a ~ so this keeps an author's name from being broken across
% two lines.
% use \thanks{} to gain access to the first footnote area
% a separate \thanks must be used for each paragraph as LaTeX2e's \thanks
% was not built to handle multiple paragraphs
%

\author{Fahem~Kebair~and~Frédéric~Serin
\thanks{F. Kebair is PhD student in computer science with LITIS--Laboratoire d'Informatique de Traitement 
de l'Information et des Systsème, University of Le Havre, 25 rue Philippe Lebon, 76058, 
Le Havre, Cedex, France, e-mail: fahem.kebair@univ-lehavre.fr}
\thanks{F. Serin is professor assistant in computer science with LITIS, e-mail: frederic.serin@univ-lehavre.fr}}

% make the title area
\maketitle

\begin{abstract}
This paper presents an agent-oriented approach to build a decision support system aimed 
at helping emergency managers to detect and to manage risks. We stress the flexibility 
and the adaptivity characteristics that are crucial to build a robust and efficient system, 
able to resolve complex problems. The system should be independent as much as possible from 
the subject of study. Thereby, an original approach based on a mechanism of perception, 
representation, characterisation and assessment is proposed. The work described here is 
applied on the RoboCupRescue application. Experimentations and results are provided.
\end{abstract}
% IEEEtran.cls defaults to using nonbold math in the Abstract.
% This preserves the distinction between vectors and scalars. However,
% if the journal you are submitting to favors bold math in the abstract,
% then you can use LaTeX's standard command \boldmath at the very start
% of the abstract to achieve this. Many IEEE journals frown on math
% in the abstract anyway.

% Note that keywords are not normally used for peerreview papers.

% For peer review papers, you can put extra information on the cover
% page as needed:
% \ifCLASSOPTIONpeerreview
% \begin{center} \bfseries EDICS Category: 3-BBND \end{center}
% \fi
%
% For peerreview papers, this IEEEtran command inserts a page break and
% creates the second title. It will be ignored for other modes.
\IEEEpeerreviewmaketitle

\section{Introduction}

The use of Decision Support Systems (DSSs) has considerably increased, during the last decade, due to the complexity of the problems faced by the decision makers. Indeed, the need for decision support tools should be, if anything, increasing \cite{shawn97}. In some domains or circumstances, making a decision is an arduous task that requires some abilities exceeding the human capacities. We can think decision-making in Simon's decision making model, which consists in intelligence, design and choice \cite{simon77}. Based on this model, the complexity of decision making lies in the difficulty to get a clear insight into the problem to resolve, to process the vast amount of collected information, to make the right choice in time and to harmonise finally the set of decisions made by the decision makers or the organisations. Therefore, computer-based systems may be very helpful to support decision making, especially when the environment problem is complex, dynamic and partially known. Processing and managing information issued from such an environment represents a challenge to the DSS developers. However, DSS are well known to be customized for a specific purpose and can rarely be reused. Moreover, DSSs only support circumstances which lie in the known and knowable spaces and do not support complex situations sufficiently \cite{french05}. This led us to think DSSs must be flexible and adaptive to be effective in solving complex problems as the risk and crisis management. Flexibility allows the use of the system in different subject of studies with minor changes. In other words, the system operates in a generic manner and relies on specific knowledge that are defined by experts of the domain. Adaptivity is an essential characteristic to build intelligent information systems which draws increasingly the attention of the scientists in computer science and in artificial intelligence. Thanks to the adaptivity, the system may adapt its behaviour autonomously by altering its internal structure and changing its behaviour to better respond to the change of its environment. The multiagent systems technology is an appropriate solution to achieve these two objectives. Intelligent agents \cite{wooldridge02} are able to self-perform actions and to interact with other agents and their environment in order to carry out some objectives and to react to changes they perceive by adapting their behaviours. 

In this paper we propose an agent-oriented approach aimed at building a DSS that has as role to help emergency managers to detect and to manage risks in emergency situations. The system perceives facts occurred in the environment, represents them and analyses them to assess the current situation. To evaluate the situation, the system uses an analogical reasoning based on the following postulate: if a given situation A seems like a situation B, then it is likely that the consequences of the situation A will be similar to those of B. Consequently, the risk appeared in B become a potential risk of A. An internal multi-level kernel is used to insure the whole decision-support process. We utilise an earthquake scenario using the RoboCupRescue Simulation System (RCRSS) \cite{kitano99} \cite{RoboCupRescue} in order to illustrate our approach. Experimentations and results are provided and discussed. 

\section{Decision Support System for Risk Detection and Management}

\subsection{Definitions and Approaches}

The Risk is a concept that denotes a potential negative impact to an asset or some characteristic of value that may arise from some present process or future event. There are many more and less precise definitions of risk. They do depend on specific applications and situational contexts. It can be assessed qualitatively or quantitatively. In our context, we are interested in natural and technological risks. The management of these risks often represented a large-scale challenge for the individuals and the organisations, since they are hard to predict and their occurrences are much sudden. The risk management may be defined as the systematic application of management policies, procedures and practices to the tasks of establishing the context, identifying, analysing, evaluating, treating, monitoring and communicating risk \cite{australia04}. This process is complex and exceeds widely the human abilities. The use of the DSS in this case is indispensable. Indeed, DSSs are interactive, computer-based systems that aid users in judgement and choice activities. They provide data storage and retrieval but enhance the traditional information access and retrieval functions with support for model building and model-based reasoning. They support framing, modeling, and problem solving \cite{druzdzel00}. In the context of the risks and crisis management, the DSS must insure the following functionalities:

\begin{itemize}
 \item Evaluation of the current situation, the system must detect/recognize an abnormal event;
\item Evaluation/Prediction of the consequences, the system must assess the event by identifying the possible consequences;
\item Intervention planning, the system must help the emergency responders in planning their interventions thanks to an actions plan (or procedures) that must be the most appropriate to the situation.
\end{itemize}

\begin{figure}[h]
% \begin{changemargin}{-0.7cm}{0cm}
\centering
\includegraphics[trim = 0mm 15cm 0mm 0cm, height=2.3in]{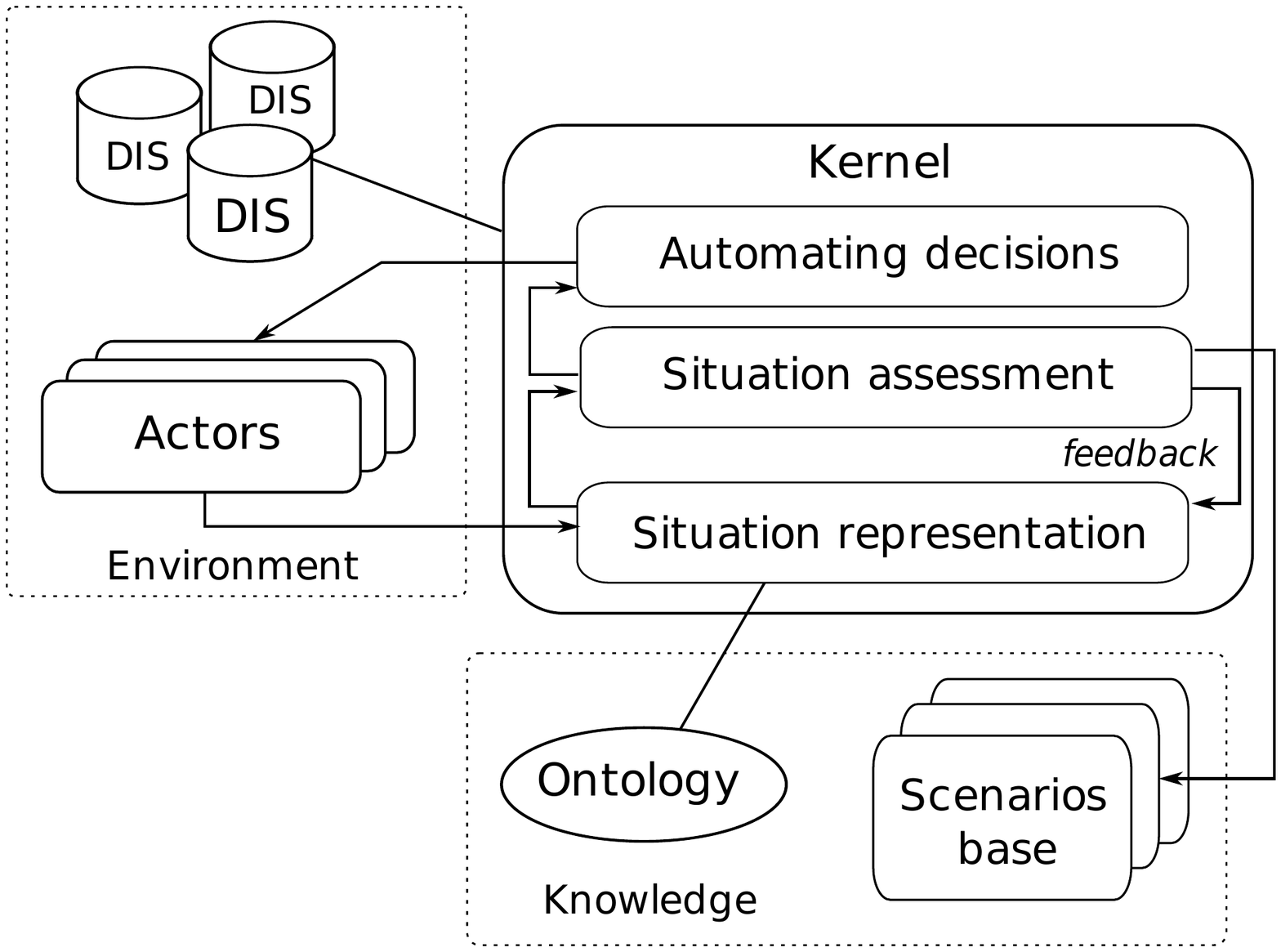}
\caption{Whole DSS architecture}
% \label{fig:dss}
% \end{changemargin}
\end{figure}

\subsection{DSS Architecture}

The kernel is the main part of the DSS and has as role to manage all the decision-support process. The environment includes essentially the actors and Distributed Information Systems (DIS) and feeds permanently the system with information describing the state of the current situation. In order to apprehend and to deal with these information, specific knowledge related to the domain as ontologies and proximity measures are required. The final goal of the DSS is to provide an evaluation of the situation by comparing it with past experimented situations stored as scenarios in a Scenario Base (SB).

The kernel is a MAS operating on three levels. It intends to detect significant organisations that give a meaning to data in order to support finally the decision making. We aim, from such a structure, to equip the system with an adaptable and a partially generic architecture that may be easily adjusted to new cases of studies. Moreover, its suppleness makes the system able to operate autonomously and to change its behaviour according to the evolution of the problem environment. As follows a description of each level:

\begin{itemize}
\item \textit{Situation representation:} One fundamental step of the system is to represent the current situation and its evolution over time. Indeed, the system perceives the facts that occur in the environment and creates its own representation of the situation thanks to a factual agents organisation. This approach has as purpose to let emerge subsets of agents. 

\item \textit{Situation assessment:} A set of assessment agents are related to scenarios stored in a SB. These agents scrutinise permanently the factual agents organisation to find agents clusters enough close to their scenarios. This mechanism is studied ``manually'' by an expert of the domain and is similar to a Case-Based Reasoning (CBR) \cite{kolodner93}, except it is dynamic and incremental. According to the application, one or more most pertinent scenarios are selected to inform decision-makers about the state of the current situation and its probable evolution, or even to generate a warning in case of detecting a risk of crisis. The evaluation of the situation will be then reinjected in the perception level in order to confirm the position of the system about the current situation. This characteristic is inspired from the feedbacks of the natural systems. In that manner, the system learns from its successes or from its failures. 

\item \textit{Automating decisions:} Outcomes generated by the assessment agents are captured by a set of performative agents and are transformed in decisions that may be used directly by the final users. 
\end{itemize}

\subsection{RoboCupRescue Case Study}

The RCRSS is an agent-based simulator which intends to reenact the rescue mission problem in real world. An earthquake scenario is reproduced including various kinds of incidents as the traffic after earthquake, buried civilians, road blockage, fire accidents, etc. A set of heterogeneous agents (RCR agents) coexist in the disaster space: rescue agents that are fire brigades, ambulance teams and police forces, and civilians agents. We focus, in this application, on the development of the rescue agents behaviours. Our final goal is to use the DSS in order to improve their decision-making ability and to support them during their rescue operations.  

A model of the RoboCupRescue disaster space and the properties of its components, and the RCR agents are detailled in \cite{takahashi01}. We use this model in order to extract knowledge and to formalise information.

\section{Dynamic Representation of the Situation: Factual Agents}

The system perceives and represents the facts occurred in the situation in an original manner using factual agents. Factual agents are reactive and proactive agents according to the agents definition given in \cite{wooldridge02}. Each agent carries an elementary datum that represents an observed fact and that aims to manage it over time. This information is presented in the shape of a Factual Semantic Feature (FSF), more details about this structure and how it is formalised and managed by a factual agent is provided in \cite{kebair08}.

The objective by using factual agents in the representation situation level is to reflect the dynamic change of the situation and to let emerge, from this view, agents subsets. These subsets may be representative of some situations that are close to some others encountered in the past. The analysis of these agents groups is based on geometric criteria, insuring thus the independence of the treatment from the subject of study. Each factual agent exposes behavioural activities that are characterised thanks to numerical indicators. The latter form a behavioural vector that draws, by its variations, the dynamics of the agent during its live. This gives a meaning to the state of the agent inside its organisation and consequently to the prominence of the semantic character that it carries. 

The goal of our approach is to characterise the factual agents organisation by forming dynamically agents clusters and comparing them with stored scenarios. The clustering algorithms seem appropriate to this objective, since they are able to create objects groups in an unsupervised way. However, these methods present some deficiencies in our case. The main ones are the need to specify some parameters as the minimal distance between two objects, required by density-based algorithms \cite{ester96}; or the minimal length of a cluster, required by Kmeans algorithms \cite{hartigan79}. Moreover, the experimentations we led using these methods showed us that we are unable to analyse instantaneously the obtained clusters neither to reproduce them. We changed therefore our way for proceeding by confiding this task to the assessment agents. These agents will search through the factual agents in order to form clusters, that should be the closest to the scenarios to which they are linked. We think this approach is more suitable for our problem, since it does not require specific knowledge and we are certain that the obtained clusters have probably a meaning and may be easily interpreted. In addition we may exploit the assets of the agents, especially their adaptivity and their communication abilities.

\section{Situation Assessment}

\subsection{Assessment Agents}
\begin{figure}
\begin{changemargin}{-0.4cm}{0cm}
% \centering
\includegraphics[trim = 0mm 16.5cm 0mm 0cm, width=0.55\textwidth]{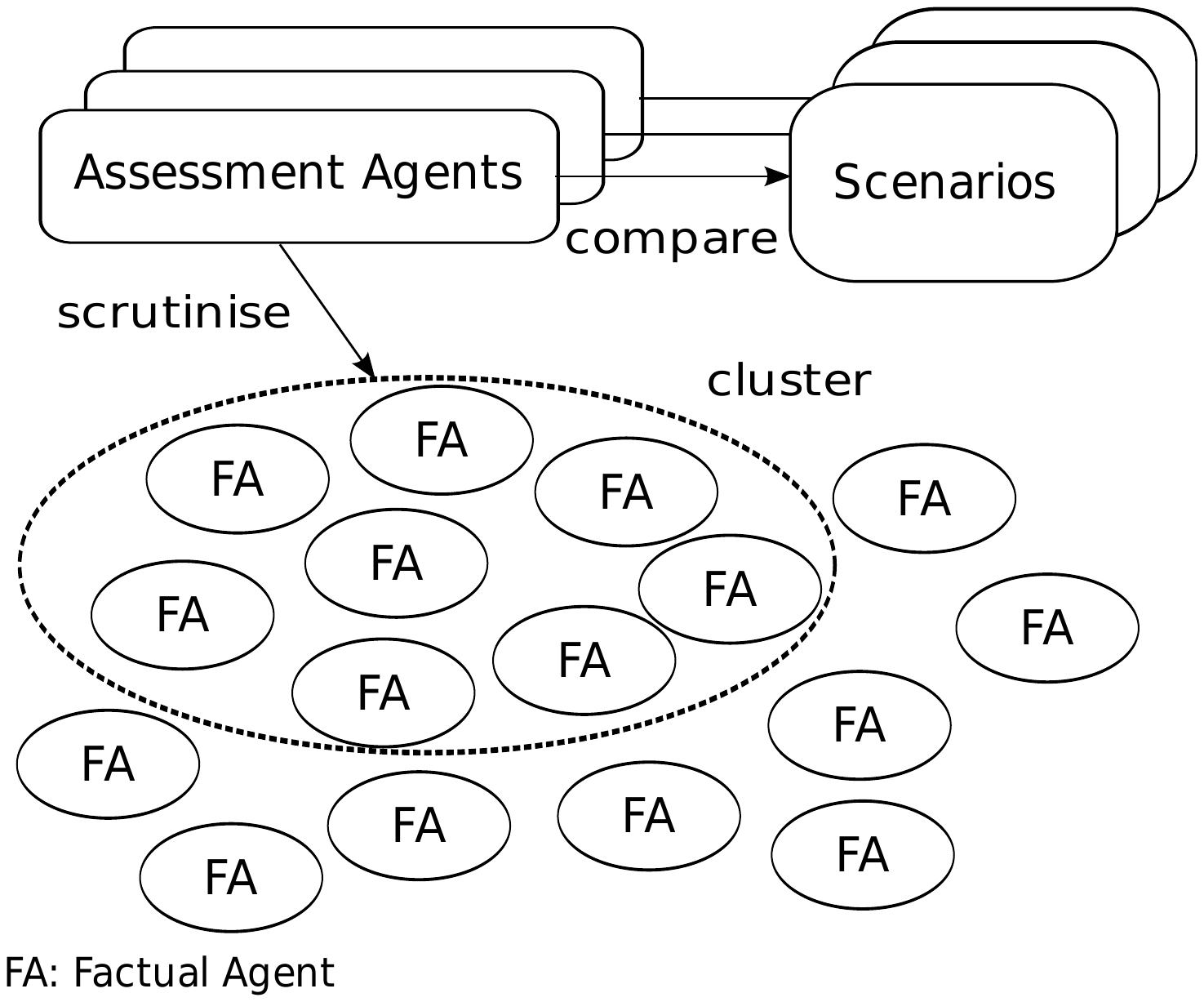}
\caption{Role of the assessment agents in the DSS}
\label{fig:level3}
\end{changemargin}
\end{figure}

Each assessment agent is linked to a scenario stored in the SB (see~\figurename\ref{fig:level3}). Each scenario is composed of one or more factual agents clusters, this depends on the treated application. A cluster is made up of a set of elements, each one includes an FSF, the indicators values of the factual agent associated to this FSF and the size of its Acquaintances Network (AN). Thus, a cluster element has the following structure: $FSF:V_{I1}\dots V_{In}:S_{AN}$, with $V_{I}$ a value of indicator $I$, and an example of an FSF is (fire, intensity, strong, location, $2^{nd}$ street, time, 10:00 pm).

The role of the assessment agents is to scrutinise permanently the organisation of the factual agents in order to extract agents clusters that should be similar as much as possible to their scenarios. A relevance, which is the sum average of all the similarities values of a created cluster elements, is attributed to each cluster to indicate its proximity to a stored scenario. This value is included in a range of [0,1]. The more the relevance is near to 1, the more the cluster is close to its scenario maker and vice versa. The clusters, and consequently the assessment agents, are sorted according to their relevances and the selected agents depend on their rank and the size of their clusters .i.e. the first agents covering the bulk of the situation are selected. 

To find close elements in the factual agents organisation, the assessment agents look only at the numeric properties of the agents and disregard the semantic characters that they carry. This insures the genericity of the mechanism. The assessment agents compare the elements of their scenarios with those carried by the factual agents by computing distances between them. The compared data are vectors defined by the $n$ indicators of the factual agent and its AN size. The cosine similarity measure (see equation~\ref{E:CS}) is used in order to compute the similarity between these vectors. The similarity value is included in a range of [0,1]. A value of 1 means the perfect equality between the two vectors, whereas 0 means their total divergence.

\begin{equation}
CS(V_{1},V_{2}) = \dfrac{ x_{1}x_{2}+y_{1}y_{2}+z_{1}z_{2}} {\sqrt{x_{1}^2+y_{1}^2+z_{1}^2}\sqrt{x_{2}^2+y_{2}^2+z_{2}^2}}
\label{E:CS}
\end{equation}

With $V_{1}$ and $V_{2}$ two vectors, and $x_{i}$, $y_{i}$ and $z_{i}$ are their respective coordinates. 

\subsection{Experimentations}

We have made experimentations on the RCR application dealing with fires situations. We have developed a prototype allowing the representation and the assessment of risks. The perceived facts in the disaster space are related to the fires propagation and to the fire brigades activities that try to extinguish these fires. The system includes a factual agents organisation for the perception and the representation of the situation and a set of assessment agents to deal with the facts evolution. At this progression stage of our work, the assessment situation is limited to the recognition of factual agents clusters according to past ones defined and experimented beforehand. We have defined therefore, from a starting scenario, a clusters set that we intend to regain in other similar scenarios by forming similar clusters. To modify an RCR scenario, we change the strategy applied by the fire brigades. This allows to have a different perception of the environment and different behaviours of the agents.

 \begin{figure*}
% \begin{changemargin}{-0.3cm}{0cm}
\centering
\includegraphics[trim = 0mm 19.5cm 0mm 0cm, width=1\textwidth]{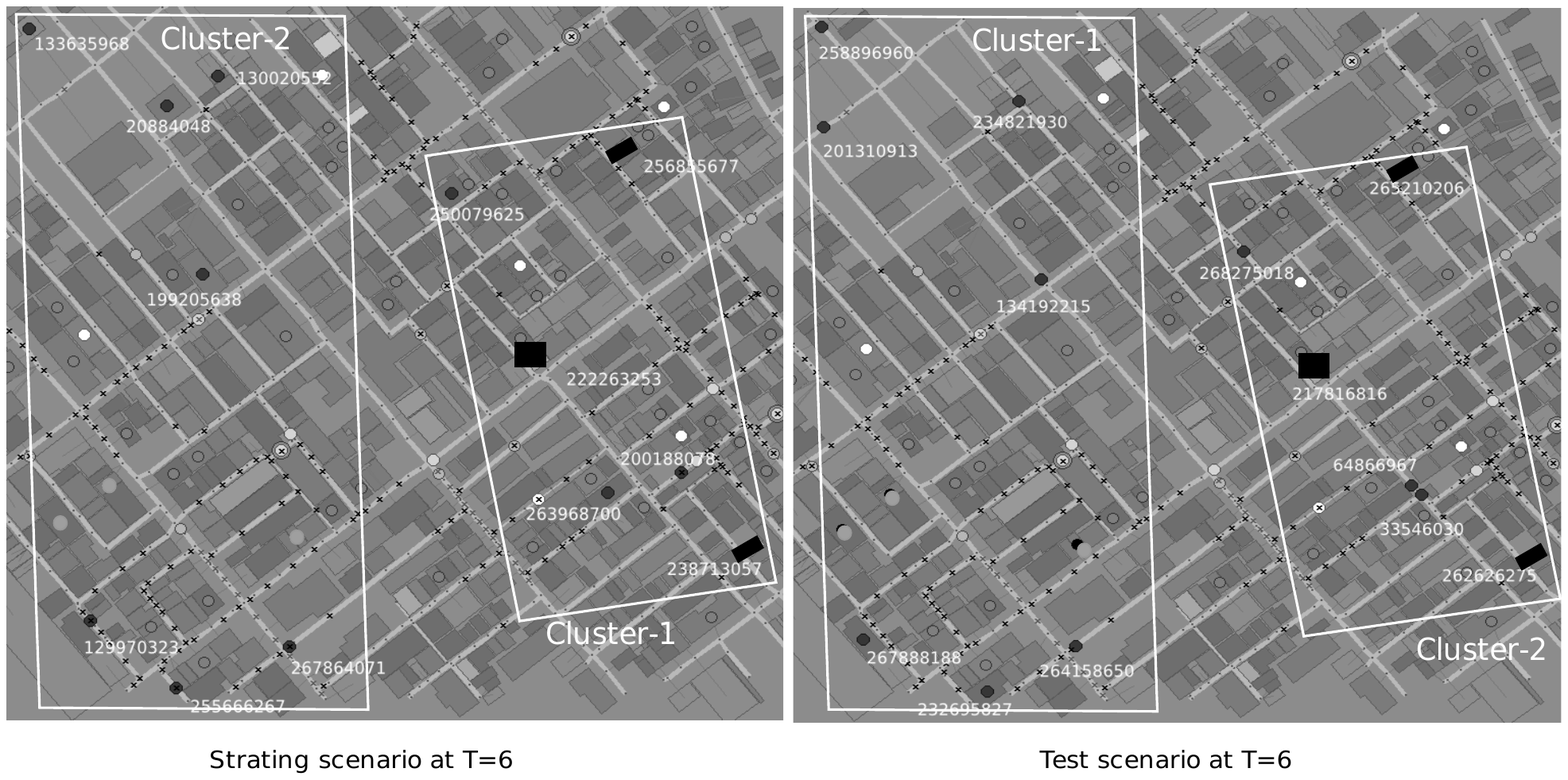}
\caption{First test example at the beginning of the RCR simulation}
\label{fig:rcr6}
% \end{changemargin}
\end{figure*}

\figurename~\ref{fig:rcr6} shows two views of the disaster space state at the beginning of the simulation--at the $6^{th}$ second. The left view belongs to the starting scenario, the right one belongs to a scenario test. What interests us in these views are the fire brigades agents represented by black ellipses and the fires represented by black rectangles. Both objects have white identifiers (IDs), we note that the RCRSS gives randomly new IDs for all the RCR objects in each new simulation. These two elements are represented in the system by two different kinds of factual agents. We have identified two factual agents clusters at this step. Cluster-1 includes starting fires and the first fire brigades having perceived these fires and which are the most able to put out them. Cluster-2 contains however the rest of the fire brigades that are in a passive state. 

Table~\ref{tab:rcr6} presents a test example. For this example we have four assessment agents, each one is associated to one cluster in the base. The table shows both the stored clusters elements and those created by the assessment agents. As we see, the two first agents (Agent-2 and Agent-1) regained two analogous clusters with relatively high relevances ($r$) in the test scenario and cover all the perceived facts of the situation. These two agents are therefore selected as the best candidates to provide the final decisions.

\begin{table}
% \begin{changemargin}{-0.1cm}{0cm}
\begin{center}
% \begin{tabular}{|p{4cm}|p{2cm}|p{4cm}|}
\caption{Created clusters at the $6^{th}$ second of the RCR simulation}
\begin{tabular}{lp{1cm}l}
\hline
% Stored clusters & Assessment agents & Similar clusters \\
\multicolumn{1}{c}{Stored clusters} & \multicolumn{1}{p{1cm}}{Assessment agents} & \multicolumn{1}{c}{Similar clusters} \\
\hline
Cluster-2:		& Agent-2 & Cluster-1, $r$=0.99\\
\hspace{0.3cm} fireBrigade\#267864071 & & \hspace{0.3cm} fireBrigade\#267888188 \\
\hspace{0.3cm} fireBrigade\#130020552 & & \hspace{0.3cm} fireBrigade\#264158650 \\
\hspace{0.3cm} fireBrigade\#129970323 & & \hspace{0.3cm} fireBrigade\#201310913 \\
\hspace{0.3cm} fireBrigade\#255666267 & & \hspace{0.3cm} fireBrigade\#134192215 \\
\hspace{0.3cm} fireBrigade\#199205638 & & \hspace{0.3cm} fireBrigade\#234821930 \\
\hspace{0.3cm} fireBrigade\#20884048 & & \hspace{0.3cm} fireBrigade\#232695827 \\
\hspace{0.3cm} fireBrigade\#133635968 & & \hspace{0.3cm} fireBrigade\#258896960 \\
\hline
Cluster-1: 		& Agent-1 & Cluster-2, $r$=0.89 \\
\hspace{0.3cm} fireBrigade\#200188078 & & \hspace{0.3cm} fireBrigade\#64866967 \\
\hspace{0.3cm} fireBrigade\#250079625 & & \hspace{0.3cm} fireBrigade\#268275018 \\
\hspace{0.3cm} fireBrigade\#263968700 & & \hspace{0.3cm} fireBrigade\#33546030 \\
\hspace{0.3cm} fire\#238713057 & & \hspace{0.3cm} fire\#265210206 \\
\hspace{0.3cm} fire\#222263253 & & \hspace{0.3cm} fire\#262626275 \\
\hspace{0.3cm} fire\#256855677 & & \hspace{0.3cm} fire\#217816816 \\
\hline
Cluster-4:		& Agent-4 & Cluster-3, $r$=0.80 \\
\hline
Cluster-3:		& Agent-3 & Cluster-4, $r$=0.67 \\
\hline
\end{tabular}
\label{tab:rcr6}
\end{center}
% \end{changemargin}
\end{table}

\begin{figure*}
% \begin{changemargin}{-0.3cm}{0cm}
\centering
\includegraphics[trim = 0mm 20.5cm 0mm 0cm, width=1\textwidth]{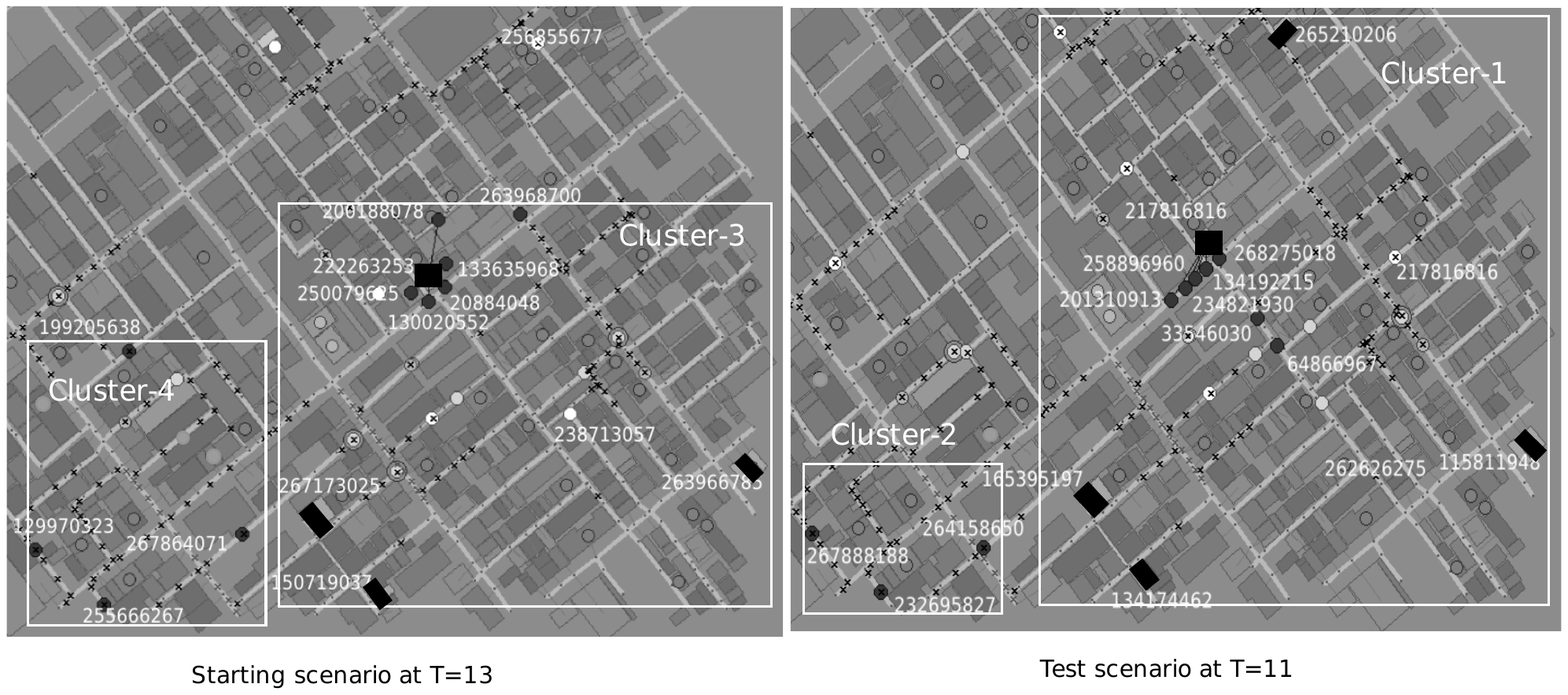}
\caption{Second test example in the middle of the RCR simulation}
\label{fig:rcr13}
% \end{changemargin}
\end{figure*}

The second example (see~\figurename~\ref{fig:rcr13}) concerns another scenario in an advanced stage of the RCR simulation--at the $13^{th}$ second of the simulation--in which fires are more important and the fire brigades are more active. At this step, two starting clusters have been identified and stored. Cluster-3 includes fire brigades in full fight with fires and other important starting fires. Cluster-4 presents some isolated fire brigades blocked by debris and that are unable to move. A similar situation is perceived at the $11^{th}$ second of the test scenario. The most relevant assessment agents are Agent-3 and Agent-4 that succeed in creating two similar clusters, whereas Agent-1 and Agent-2 have retrogressed in the relevances rank. 

\begin{table}[h]
\begin{center}
\caption{Created clusters at the $11^{th}$ second of the RCR simulation}
\begin{tabular}{lp{1cm}l}
\hline
% Stored clusters & Assessment agents & Similar clusters \\
\multicolumn{1}{c}{Stored clusters} & \multicolumn{1}{p{1cm}}{Assessment agents} & \multicolumn{1}{c}{Similar clusters} \\
\hline
Cluster-3:		& Agent-3 & Cluster-1, $r$=0.83 \\
\hspace{0.3cm} fireBrigade\#200188078 & & \hspace{0.3cm} fireBrigade\#201310913 \\
\hspace{0.3cm} fireBrigade\#263968700 & & \hspace{0.3cm} fireBrigade\#134192215 \\
\hspace{0.3cm} fireBrigade\#133635968 & & \hspace{0.3cm} fireBrigade\#234821930 \\
\hspace{0.3cm} fireBrigade\#20884048 & & \hspace{0.3cm} fireBrigade\#268275018 \\
\hspace{0.3cm} fireBrigade\#130020552 & & \hspace{0.3cm} fireBrigade\#64866967 \\
\hspace{0.3cm} fireBrigade\#250079625 & & \hspace{0.3cm} fireBrigade\#258896960 \\
\hspace{0.3cm} fire\#222263253 & & \hspace{0.3cm} fire\#265210206 \\
\hspace{0.3cm} fire\#263966785 & & \hspace{0.3cm} fire\#217816816 \\
\hspace{0.3cm} fire\#267173025 & & \hspace{0.3cm} fire\#134174462 \\
\hspace{0.3cm} fire\#150719037 & & \hspace{0.3cm} fire\#165395197 \\
			       & & \hspace{0.3cm} fire\#115811948 \\
\hline
Cluster-4: 		& Agent-4 & Cluster-2, $r$=0.80 \\
\hspace{0.3cm} fireBrigade\#199205638 & & \hspace{0.3cm} fireBrigade\#264158650 \\
\hspace{0.3cm} fireBrigade\#267864071 & & \hspace{0.3cm} fireBrigade\#267888188 \\
\hspace{0.3cm} fireBrigade\#255666267 & & \hspace{0.3cm} fireBrigade\#232695827 \\
\hspace{0.3cm} fireBrigade\#129970323 & & \\
\hline
Cluster-1		& Agent-1 & Cluster-3, $r$=0.78 \\
\hline
Cluster-2		& Agent-2 & Cluster-4, $r$=0.44 \\
\hline 
\end{tabular}
\label{tab:rcr13}
\end{center}
\end{table}

\section{Conclusion}

We have described in this paper an agent-based approach that aims to build a DSS. The system intends to help emergency planners to detect risks and to manage crisis situations by perceiving, representing and assessing a current situation. We think this approach may be adjusted easilly to different problems types and enables the system to have an adaptive behaviour thanks to a multiagent multilevel kernel. We are working currently on the assessment level of the system mechanism. We have presented here first results applied on the RoboCupRescue. We intend to apply this approach on different subjects of studies in order to better improve its generic aspect. We aim also to generalise this approach by setting up a generic modelling of factual agents clusters that will enhance their formalisation and their management.

% \begin{figure*}[!t]
% \centerline{\subfloat[Case I]\includegraphics[width=2.5in]{system.pdf}%
% \label{fig_first_case}}
% \hfil
% \subfloat[Case II]{\includegraphics[width=2.5in]{subfigcase2}%
% \label{fig_second_case}}}
% \caption{Simulation results}
% \label{fig_sim}
% \end{figure*}

%\begin{table}[!t]
%% increase table row spacing, adjust to taste
%\renewcommand{\arraystretch}{1.3}
% if using array.sty, it might be a good idea to tweak the value of
% \extrarowheight as needed to properly center the text within the cells
%\caption{An Example of a Table}
%\label{table_example}
%\centering
%% Some packages, such as MDW tools, offer better commands for making tables
%% than the plain LaTeX2e tabular which is used here.
%\begin{tabular}{|c||c|}
%\hline
%One & Two\\
%\hline
%Three & Four\\
%\hline
%\end{tabular}
%\end{table}

\end{document}